\documentclass[accept]{elsarticle}

\usepackage{epsfig}
\usepackage{graphicx}
\usepackage{amsmath}
\usepackage{amssymb}
\usepackage{algorithm}
\usepackage{algpseudocode}
\usepackage{subfigure}
\usepackage{multirow}
\usepackage{color}
\usepackage{flushend}
\usepackage{soul}
\soulregister\cite7
\soulregister\ref7

\usepackage{lineno,hyperref}

\newcommand{\etal}{\textit{et al}.}
\newcommand{\ie}{\textit{i}.\textit{e}.}
\soulregister\etal7

\begin{document}

\begin{frontmatter}

\title{SOSD-Net: Joint Semantic Object Segmentation and Depth Estimation from Monocular images}

\author[mymainaddress,mysecondaryaddress]{Lei He}
\ead{helei07@baidu.com}

\author[mysecondaryaddress]{Jiwen Lu}
\ead{lujiwen@tsinghua.edu.cn}

\author[mythirdaddress]{Guanghui Wang}
\ead{wangcs@ryerson.ca}

\author[mymainaddress]{Shiyu Song}
\ead{songshiyu@baidu.com}

\author[mysecondaryaddress,myfourthaddress]{Jie Zhou\corref{mycorrespondingauthor}}
\ead{jzhou@tsinghua.edu.cn}
\cortext[mycorrespondingauthor]{Corresponding author}

\address[mymainaddress]{Baidu Autonomous Driving Technology Department (ADT)}
\address[mysecondaryaddress]{Beijing National Research Center for Information Science and Technology (BNRist), Department of Automation, Tsinghua University, Beijing, 100084, China}
\address[mythirdaddress]{Department of Computer Science, Ryerson University, Toronto, ON, Canada M5B 2K3}
\address[myfourthaddress]{Tsinghua Shenzhen International Graduate School, Tsinghua University, Shenzhen, 518055, China}

\begin{abstract}
Depth estimation and semantic segmentation play essential roles in scene understanding. The state-of-the-art methods employ multi-task learning to simultaneously learn models for these two tasks at the pixel-wise level. They usually focus on sharing the common features or stitching feature maps from the corresponding branches. However, these methods lack in-depth consideration on the correlation of the geometric cues and the scene parsing. In this paper, we first introduce the concept of semantic objectness to exploit the geometric relationship of these two tasks through an analysis of the imaging process, then propose a Semantic Object Segmentation and Depth Estimation Network (SOSD-Net) based on the objectness assumption. To the best of our knowledge, SOSD-Net is the first network that exploits the geometry constraint for simultaneous monocular depth estimation and semantic segmentation. In addition, considering the mutual implicit relationship between these two tasks, we exploit the iterative idea from the expectation-maximization algorithm to train the proposed network more effectively. Extensive experimental results on the Cityscapes and NYU v2 dataset are presented to demonstrate the superior performance of the proposed approach.
\end{abstract}


\end{frontmatter}


\section{Introduction}

Depth estimation and semantic segmentation, as two major components in scene understanding, have received a lot of attention in the computer vision community. In recent years, with the successful applications of deep convolutional neural networks, the performance of depth estimation and semantic segmentation has been greatly improved~\cite{long2015fully,badrinarayanan2017segnet,he2016fast,eigen2015predicting,fu2018deep}, owing to the superior representation ability of the deep features over the classical handcrafted features~\cite{cen2020deep,karsch2014depth,ma2020mdfn,liu2018semantic,yan2019triplet}.

Monocular depth estimation is an essential approach in understanding the 3D geometry of a scene~\cite{eigen2014depth,eigen2015predicting,he2018learning,fu2018deep,yang2015semantic}. Depth estimation is usually formulated as a regression problem that assigns each pixel a continuous depth value. However, this task exists inherent ambiguity with some scene priors, as analyzed in He~\etal~\cite{he2018learning}. The scene priors refer to the elements that can remedy the ambiguity of the monocular depth estimation, such as the physical size of the objects in the scene, and the focal length information of the camera, etc. To improve the accuracy of monocular depth estimation, these ambiguous elements need to be properly integrated into the network during the training and inferring process of the network. With the multi-scale fusion and hierarchical representation of deep networks, the precision of semantic segmentation~\cite{long2015fully,badrinarayanan2017segnet,chen2018encoder} has been greatly improved. Nevertheless, most segmentation models have limitations in certain scenarios, like segmenting slender objects such as poles. If we can obtain an accurate depth map, there generally exists a depth margin between the poles and the surrounding background or objects. Thus, the depth information can greatly help to improve the segmentation performance, especially in challenging situations.

In order to explore the correlation of the depth information and semantic segmentation, jointly train a network to simultaneously learn the two tasks become an attractive direction in scene understanding. One popular approach is to combining multi-task neural activations via employing the network architecture interaction~\cite{wang2015towards,misra2016cross,zhang2018joint}. However, the geometric constraint is not explicitly explored in the fusion process. To obtain an optimal descent direction of the common weights, another approach~\cite{kendall2018multi,chen2018gradnorm,sener2018multi} is designing the joint-optimization objective functions by adaptively selecting loss weight of each task during the training phase. This approach is designed only to pursue a better sharing feature representation, without considering the geometric relationship of the two tasks.

\begin{figure}
\begin{center}
   \includegraphics[width=0.95\linewidth]{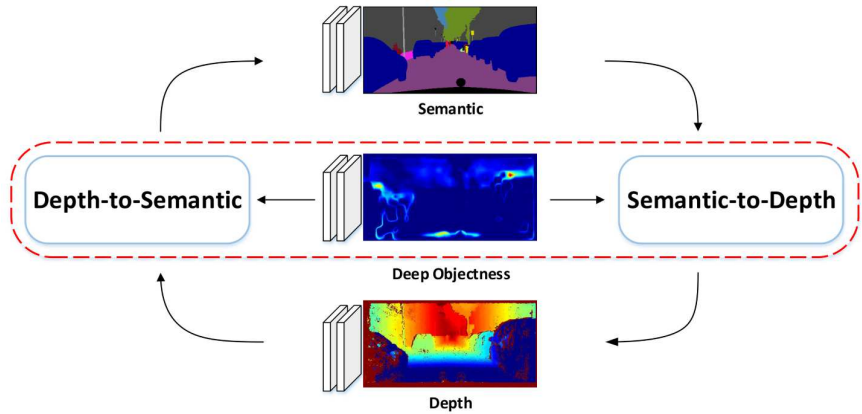}
\end{center}
   \caption{Joint optimization of monocular depth, semantic, and semantic objectness.}
\label{fig:dosnet}
\end{figure}

In this paper, we propose to explore the geometric relationship between monocular semantic segmentation and depth estimation, and design a novel neural network (SOSD-Net) to embed the semantic objectness, making it possible to simultaneously learn the geometric cues and scene parsing, as shown in Figure~\ref{fig:dosnet}. The proposed network is designed strictly according to the geometric constraints to boost up the performance of the two tasks by integrating the information of the objectness.

Specifically, when inferring the monocular depth information, the semantic objectness will fuse the features from the semantic segmentation, and vice versa. In addition, the supervised learning of the two tasks is essentially a parameter estimation problem of a Gaussian mixture model. Inspired by the idea of Expectation Maximization (EM) algorithm, we propose an effective learning strategy to alternative optimize the weights of the scene parsing and the geometric cues during the training phase. The proposed method is extensively evaluated on the CityScapes~\cite{cordts2016cityscapes} and NYU v2 datasets~\cite{silberman2012indoor}, and the experimental validation shows that the SOSD-Net outperforms the state-of-the-art multi-task approaches in the one-stage training phase, demonstrating the effectiveness of our proposed algorithm.

In summary, the key contributions of this paper include:
\begin{enumerate}
\setlength{\itemsep}{-2pt}
  \item We propose a Semantic Objectness Segmentation and Depth Estimation Network (SOSD-Net) to enhance the learning ability of joint monocular depth estimation and semantic segmentation.
  \item An effective learning strategy is proposed to alternatively update the specific weights of SOSD-Net, which significantly improves the performance of the two tasks.
  \item We achieve competing results over the state-of-the-art one-stage models on two popular benchmarks.
\end{enumerate}

\section{Related Work}

In this section, we review the related work in the following three problems: semantic segmentation, depth estimation, and multi-task learning.

\subsection{Semantic Segmentation}

With the powerful representational and inferring ability, many models~\cite{sermanet2013overfeat,sun2017revisiting} based on the deep convolutional neural networks have achieved significant improvement on several segmentation datasets, especially compared with the classical hand-crafted methods. Long~\etal~\cite{long2015fully} made a breakthrough by successfully converting the classification network to a pixel-wise segmentation network, replacing the fully connected layers with convolutional layers. Inspired by this idea, the recent semantic segmentation networks can be broadly classified into three categories. The first group~\cite{ronneberger2015u,lin2017refinenet,badrinarayanan2017segnet,noh2015learning,pohlen2017full} designs convolutional encoder-decoder network structures to gradually capture the high semantic information and recovering the spatial information. The second group of methods~\cite{liu2015parsenet,zhao2017pyramid,chen2017rethinking,chen2018deeplab,chen2018encoder} is to exploit multi-scale information to grasp better global and contextual information. The last group~\cite{chen2014semantic,zheng2015conditional,liu2015semantic,lin2016efficient,jampani2016learning,chen2018deeplab} is to explore the conditional Markov Random Field to optimize the segmentation result. In addition, Kre{\v{s}}o~\etal~\cite{krevso2016convolutional} propose a novel scale selection layer to extract convolutional features at the scale of the reconstructed depth to improve the performance of the semantic segmentation.

\subsection{Depth Estimation}
Learning depth from a single image has been extensively studied in the literature. To tackle this task, classic methods~\cite{hoiem2005automatic,saxena20083,saxena2009make3d} usually make strong geometric assumptions about the scene structure, and employ the Markov Random Field (MRF) to infer the depth by leveraging the hand-crafted features. Non-parametric algorithm~\cite{konrad2013learning,karsch2014depth} is another type of classical methods, which employ global scene features to search for candidate images that are close to the input image from a training database in the feature space. Other methods are based on the advanced deep learning models~\cite{eigen2014depth,eigen2015predicting,liu2016learning,laina2016deeper,xu2017multi,fu2018deep,He2018SpindleNetCF,liu2020depth}. Eigen~\etal~\cite{eigen2014depth} addressed this issue by fusing the depths from the global network and refined network, which was extended to use a multi-scale convolutional network in a deeper neural network~\cite{eigen2015predicting}. Recently, the unsupervised learning methods~\cite{garg2016unsupervised,godard2017unsupervised,zhou2017unsupervised,mahjourian2018unsupervised,fan2020deep} achieved significant progress. By exploiting the epipolar geometry constraints,~\cite{garg2016unsupervised,godard2017unsupervised,zhou2017unsupervised,mahjourian2018unsupervised} taken the inferred monocular depth as an intermediate result in computing the reconstruction loss. Due to the inherent ambiguity of the monocular depth estimation, He~\etal~\cite{he2018learning} proposed a novel deep neural network to remedy the ambiguity caused by the focal length.

\subsection{Multi-task Learning}

Multi-task learning aims to improve the performance of various computer vision problems. According to the design of network structure and loss function, the methods of multi-task learning are mainly divided into two categories. One of the methods~\cite{wang2015towards,mousavian2016joint,misra2016cross,zhang2018joint} is to let the network automatically learn the connection relationship among tasks, where the loss function is a weighted sum of all branches. Another method~\cite{kendall2018multi,chen2018gradnorm,sener2018multi} is to search the optimal descent direction of the gradient by adaptively selecting the weighting factors during the training process. Xu~\etal~\cite{xu2018pad} proposed a PAD-Net to utilize auxiliary tasks to facilitate optimizing the semantic segmentation and depth estimation. R. Zamir~\etal~\cite{zamir2018taskonomy} proposes a fully computational approach to model the structure of space of visual tasks, building the relationship between the depth estimation and the semantic segmentation from normals. However, these methods only directly learn the two tasks without explicitly exploring the geometric constraints between the monocular depth estimation and semantic segmentation. In this work, we propose an SOSD-Net to achieve a deep geometric relationship between monocular depth estimation and semantic segmentation.

\begin{figure}
\begin{center}
   \includegraphics[width=0.6\linewidth]{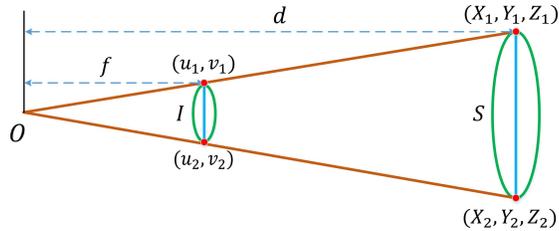}
\end{center}
   \caption{The projection process of a planar object. Where $O$ is the optical center, $I$ is the image of the planar object $S$, $d$ is the depth of the object, and $f$ is the focal length.}
\label{fig:projective}
\end{figure}

\section{Method}

\label{sec:method}
In this section, we describe the proposed SOSD-Net for monocular depth estimation and semantic segmentation. We first introduce the geometry constraint to embed the deep relation of the monocular depth and semantic information, then elaborate the network architecture of the SOSD-Net. Finally, we present the details of the proposed learning strategy.

\subsection{Geometry Constraint}
Without loss of generality, we assume the space object is linear, as shown in Figure~\ref{fig:projective}. According to the perspective projection model \cite{wang2011guide}, the image of the planar space object $S$ under $(f,O)$ is $I$, which can be formulated by the following equation.

\begin{equation}\label{eq:projective_cimg}
d\left[ {\begin{array}{c}
   u_{1} \\
   v_{1} \\
   1 \\
\end{array} } \right] = \left[ {\begin{array}{ccc}
   f_{x} & 0 & u_{x} \\
   0 & f_{y} & u_{y} \\
   0 & 0 & 1 \\
\end{array} } \right]\left[ {\begin{array}{c}
   X_{1} \\
   Y_{1} \\
   d \\
\end{array} } \right]
\end{equation}
where ($X_{1}, Y_{1}$) is the coordinate of a space point, $(u, v)$ is the coordinates of the space point on the image, ($u_{x}$, $u_{y}$) is the coordinate of the principal point, and ($f_{x}, f_{y}$) corresponds to the camera's focal length.

Through the above projection equation, it is notable that the monocular depth estimation is an ill-posed problem, which makes it difficult to accurately recover the true depth. However, if we only consider the object-level depth and assume that the depth of the inner region of the object is approximately consistent, the geometric relationship can be reduced to the following 2D-3D size information of an object.

\begin{equation}\label{eq:semantic_depth}
    \Delta u = \frac{f_{x}\Delta X}{d},\;
   \Delta v = \frac{f_{y}\Delta Y}{d}
\end{equation}
where
$\Delta u = u_{1} - u_{2},\; \Delta v = v_{1} - v_{2}$,
$\Delta X = X_{1} - X_{2},\; \Delta Y = Y_{1} - Y_{2}$.
Furthermore, we can extend the above 2D-3D size information to 2D-3D area information as below.

\begin{equation}\label{eq:area}
    d^{2} = \frac{f_{x}f_{y}\Delta X\Delta Y}{\Delta u\Delta v}
\end{equation}

The geometric relationship of the equation (3) is called  semantic objectness in this paper, which embeds the correlation of the semantic and the corresponding depth. In general, after semantic segmentation, we can obtain the 2D area information $\Delta u \Delta v$ of an object by implementing a simple post-process operation. In addition, the area information $\Delta X \Delta Y$ of an object under a specific perspective is unique. Thus, we can establish a close relationship between the object-level semantic and the corresponding depth.

In practice, the depth of the inner region of most objects is not consistent. However, if only taking a local area of the object into consideration, then the assumption of consistent depth is satisfied. Current deep neural networks can express very complex functions due to their non-linearity and a large number of parameters. Therefore, we use this powerful tool to express the local implicit relationship, and introduce a novel deep convolutional network (SOSD-Net) to embed the semantic objectness relation between the monocular depth estimation and semantic segmentation.

\begin{figure}
\begin{center}
   \includegraphics[width=1\linewidth]{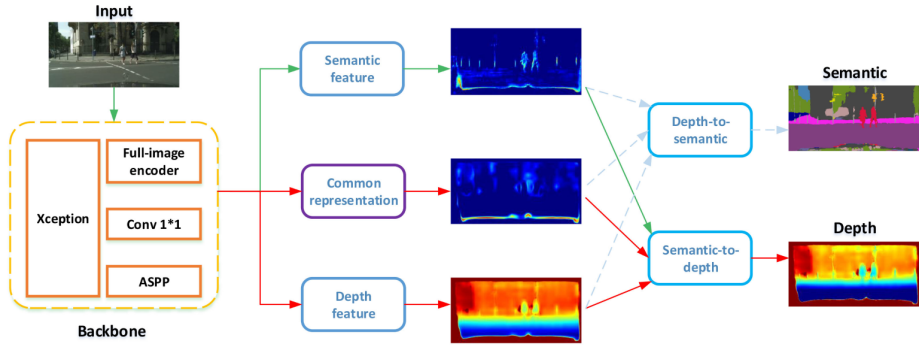}
\end{center}
   \caption{Our proposed SOSD-Net architecture leverages a shared-encoder backbone and a Decoder for semantic feature, common representation and depth feature, followed by depth-to-semantic and semantic-to-depth modules to learn semantic segmentation and depth estimation from a single image, respectively.}
\label{fig:network}
\end{figure}

\subsection{SOSD-Net Architecture}
The overall SOSD-Net architecture is depicted in Figure~\ref{fig:network}. It consists of four components as described below: a CNN backbone to extract a contextual feature, a Decoder  for three feature maps (Common Representation, Semantic Feature, Depth Feature), a semantic-to-depth unit to learn monocular depth, and a depth-to-semantic unit to learn semantic segmentation.
\begin{figure}
	\begin{center}
		\includegraphics[width=1.0\linewidth]{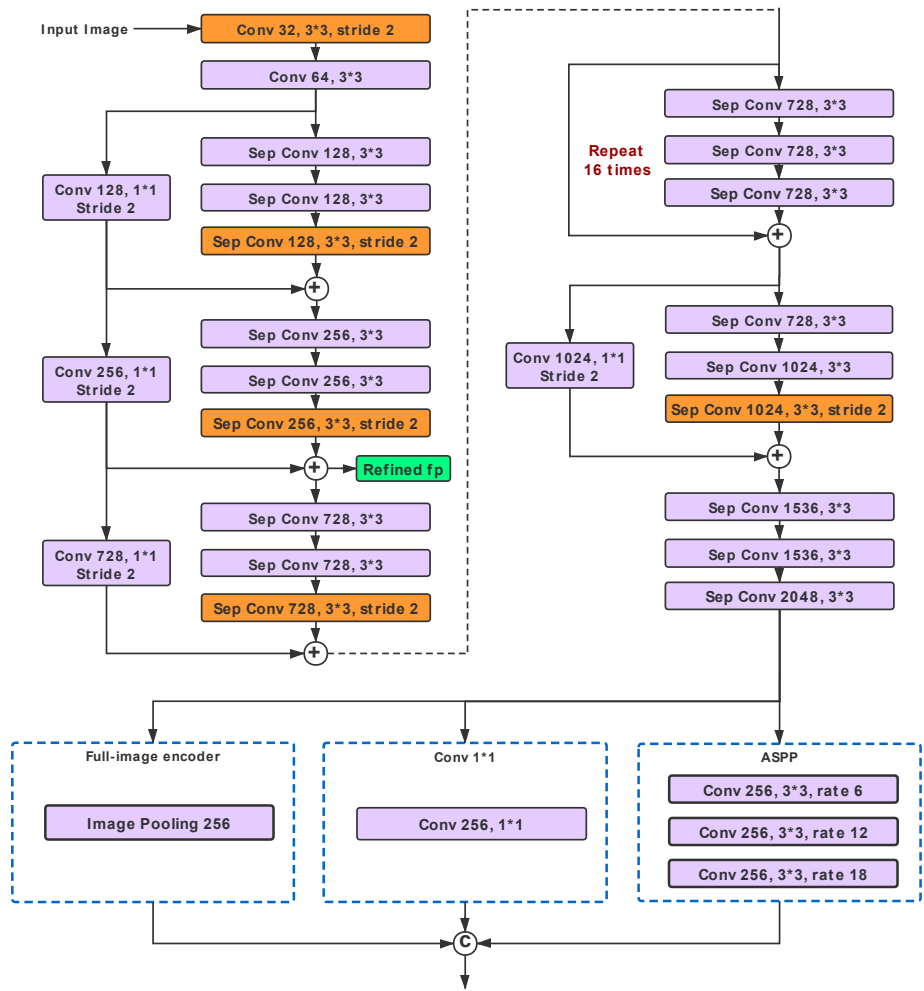}
	\end{center}
	\caption{The detailed structure of the Backbone.}
	\label{fig:network-backbone}
\end{figure}

\subsubsection{Backbone}
The backbone takes an input image and generates an intermediate feature map to be processed by each subtask. Similar to DeepLabV3+~\cite{chen2018encoder}, the backbone of the proposed SOSD-Net consists of xception-65~\cite{chollet2017xception} and three parallel components,~\ie, an atrous spatial pyramid pooling (ASPP), a cross-channel learner, and a full-image extractor, as shown in Figure~\ref{fig:network-backbone}. The ASPP and the pure $1\times1$ convolution are applied to effectively fuse complex contextual information, guided by the global information from the full-image extractor. 

\begin{figure}
	\begin{center}
		\includegraphics[width=1.0\linewidth]{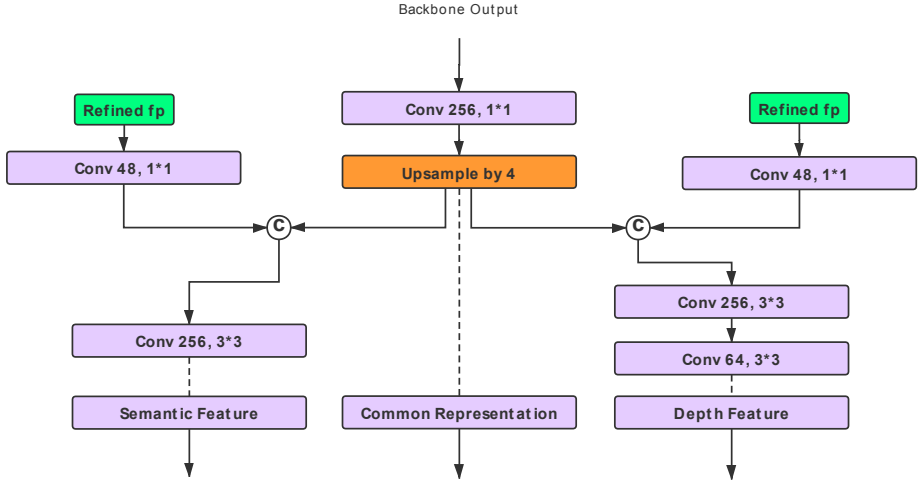}
	\end{center}
	\caption{The detailed structure of the Decoder. The Refined fp (green block) is generated from the Backbone.}
	\label{fig:network-decoder}
\end{figure}

\subsubsection{Decoder}
Based on the global feature map from the Backbone, the role of the decoder is mainly to extract fine-grained feature maps for Semantic Feature, Common Representation, and Depth Feature, respectively. In order to remedy the structure loss caused by the stride convolution, the decoder fuses the Refined fp (green block) from the Backbone in Figure~\ref{fig:network}. Based on the Refined fp, we first take one convolution layer to extract information for each task, respectively. Then combining the upsampling global feature maps, the decoder employs one convolutional layer and two convolutional layers to generate the Semantic Feature and Depth Feature, respectively. The detailed parameters of the decoder are shown in Figure~\ref{fig:network-decoder}.

\begin{figure}
	\begin{center}
		\includegraphics[width=0.8\linewidth]{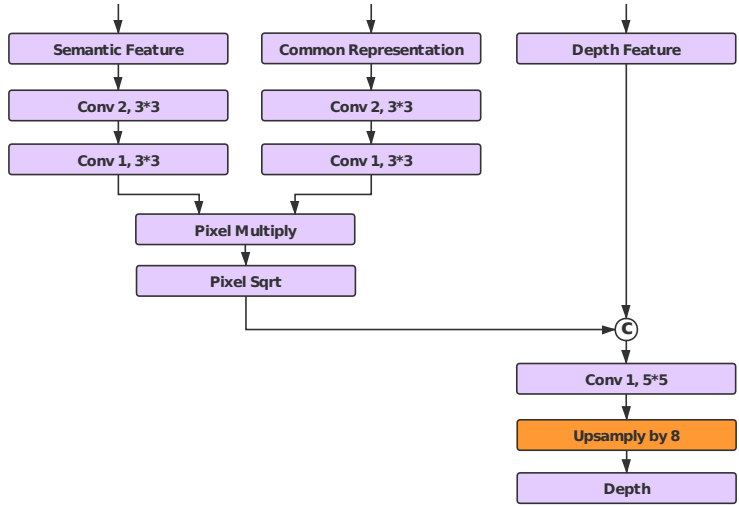}
	\end{center}
	\caption{The structure of the semantic-to-depth module.}
	\label{fig:network-semantic-to-depth}
\end{figure}

\subsubsection{Semantic-to-Depth}
Having obtained the common features, a classic decoder for monocular depth estimation employs the skip-connection and upsampling modules to obtain the high-resolution depth maps. The weights of the network are updated by minimizing the depth loss function. In order to embed the semantic objectness information for the monocular depth estimation, as described in equation~\eqref{eq:area}, we propose a semantic-to-depth module to effectively fuse the deep 2D-3D area information. As shown in Figure~\ref{fig:network-semantic-to-depth}, the deep 3D area information is extracted from the common feature maps, defined as the deep area information of an object under a certain perspective.

In order to maintain the detailed structure of the subtasks, we set the stride of the convolution to 1. In the semantic-to-depth unit of the SOSD-Net, we first utilize two convolution layers with 2 and 1 channels to generate a heatmap, which is referred to as 3D latent shared representation of an object, related to $\Delta X \Delta Y$, as shown in Figure~\ref{fig:network-semantic-to-depth}. In addition, we take another two convolution layers with 2 and 1 channels to obtain another heatmap from the semantic segmentation, which is the 2D latent shared representation of an object, related to $(\Delta u \Delta v)^{-1}$. Since the public datasets are of fixed-focal-length, we use batch normalization to automatically embed the focal length information into the two sub-branches. Next, we employ the deep 2D-3D area feature to infer the depth cue by leveraging pixel-wise multiplication and square root operations. After combining the information from the depth features and the Pixel Sqrt, this module conducts a convolution layer with 1 channel to infer depth maps. At last, we employ an upsampling operation (bilinear interpolation) on the previous depth to obtain the full-resolution depth.

\begin{figure}
	\begin{center}
		\includegraphics[width=0.8\linewidth]{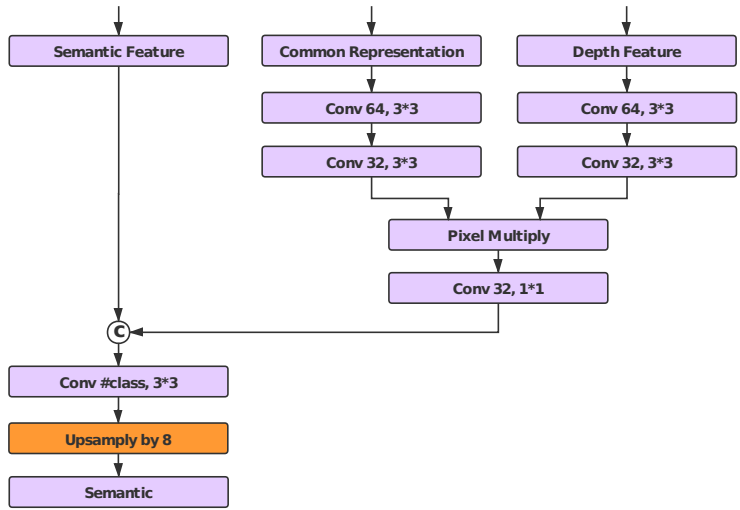}
	\end{center}
	\caption{The structure of the depth-to-semantic module.}
	\label{fig:network-depth-to-semantic}
\end{figure}

\subsubsection{Depth-to-Semantic}
Similar to the semantic-to-depth, the semantic branch also embeds the deep 2D-3D area feature by integrating the features from previous components. However, in terms of implementation details, the depth-to-semantic is different from the semantic-to-depth, as shown in Figure~\ref{fig:network-depth-to-semantic}. For example, we first apply two convolution layers with 64 and 32 channels to generate a latent variable from the pure depth branch, which is related to the $d^{-2}$. As for the 3D latent shared representation of an object, we realize it by leveraging another two convolution layers with 64 and 32 channels on the common representation. After fusing the information from the two sub-branches by pixel-wise multiplication, we employ a $1 \times 1$ convolution with 32 channels to parse the semantic cue. Having concatenated the feature maps from the semantic feature and the semantic cue, this module adds one convolution layer  to infer the semantic segmentation. To obtain the full-resolution segmentation, we take the same upsampling operation to increase the resolution of the semantic segmentation.

\begin{algorithm}[tb]
\caption{EM Learning Strategy}
\label{alg:eml}
\begin{algorithmic}[1]
\State Parameters Initialization, set $p \to 0$
\For{$i=1$ to $N$}
    \If {$ p = 0$} \Comment{learning depth}
    \For{$t=1$ to $3$} \Comment{$(\theta^{dep}, \theta^{3d}, \theta^{2d})$}
    \State $\Theta(t) = \Theta(t) - \eta\bigtriangledown_{\Theta(t)}\varphi(I, x_{sem};\Theta)$
    \EndFor
    \State $\theta^{com} = \theta^{com} - \eta\sum_{t=1}^{3}\alpha^{t}\bigtriangledown_{\theta^{com}}\varphi(I, x_{sem};\Theta)$
    \State $p\to 1$
    \Else \Comment{learning semantic}
    \For{$t=1$ to $3$} \Comment{$(\theta^{sem}, \theta^{3d}, \theta^{d^{-2}})$}
    \State $\Theta(t) = \Theta(t) - \eta\bigtriangledown_{\Theta(t)}\varphi(I, x_{dep};\Theta)$
    \EndFor
    \State $\theta^{com} = \theta^{com} - \eta\sum_{t=1}^{3}\alpha^{t}\bigtriangledown_{\theta^{com}}\varphi(I, x_{dep};\Theta)$
    \State $p\to 0$
    \EndIf
\EndFor
\end{algorithmic}
\end{algorithm}

\subsection{Learning and Loss Function}
In essence, the weight learning of deep networks is equivalent to a problem of maximum likelihood estimation. In the field of classical machine learning, simultaneously learning the depth and semantic segmentation from a single image can be regarded as a Gaussian Mixture Model (GMM), which can be effectively solved by the EM algorithm~\cite{dempster1977maximum}. In the process of parameter optimization, EM can simplify the complex estimation problem. It first optimizes some parameters ($\phi_{1}$) by fixing other parameters ($\phi_{2}$) in the parameter space, and then optimize the other parameters $\phi_{2}$ by fixing the parameters $\phi_{1}$ until achieving the optimal parameters. Inspired by the strategy of EM, we propose an effective training method to alternatively learn the weights of the SOSD-Net by taking the deep 2D-3D area information as hidden variables, which first learns the weight of the depth branch by fixing the weight of the semantic branch, and then learns the weight of the semantic branch by fixing the weight of the depth branch until the convergence of the proposed model.

Let $Y=\varphi(I, x_{sem}; \Theta)$ denote the fused outputs of the depth branch given an image $I$ and its semantic feature $x_{sem}$, where $\Theta = (\theta^{dep}, \theta^{3d}, \theta^{2d}, \theta^{com})$ corresponds to the parameters involved in the depth features, $\Delta X \Delta Y$, $(\Delta x \Delta y)^{-1}$ and backbone network, respectively, as shown in Figure~\ref{fig:network-semantic-to-depth}. As for the learning process of the monocular depth, we first update the weights of the semantic-to-depth, and then merge the backward loss from each branch to learn the weights of the common backbone. For example, for the green branch ($(\Delta x \Delta y)^{-1}$) in Figure~\ref{fig:network-semantic-to-depth}, the update process of the weights can be formulated by the following equation.
\begin{equation}\label{eq:em_each_branch}
    \theta^{2d}(t) = \theta^{2d}(t) - \eta\bigtriangledown_{\theta^{2d}(t)}\varphi(I,x_{sem};\Theta)
\end{equation}
where $\eta$ is the learning rate, and $\bigtriangledown_{\theta^{2d}(t)}\varphi(I,x_{sem};\Theta)$ is the gradient of $\varphi$ with respect to $\theta^{2d}$.

We use the same strategy of weights update for the purple branch, and the orange branch, respectively. Having updated the weights of the three branches, we merge the backward loss and learn the weights of the backbone as the following equation.
\begin{equation}\label{eq:em_common}
    \theta^{com} = \theta^{com} - \eta\sum_{t=1}^{3}\alpha^{t}\bigtriangledown_{\theta^{com}}\varphi(I, x_{sem};\Theta)
\end{equation}
where $\alpha^{t}$ is the weighting factor of the t-subnet in terms of gradient-based backpropagation. We set $\alpha^{t}$ to 1 in this paper.

Similarly, when learning the semantic segmentation, we use the same strategy to lean the weights of the proposed model, taking the monocular depth information and the deep 3D area information as hidden variables. The detailed learning strategy is shown in Algorithm~\ref{alg:eml}. Finally, the parameters of the proposed network are learned by the gradient-based backpropagation method, whose goal is minimizing the loss function defined on the prediction and the ground truth.

\textbf{Semantic segmentation.} The cross-entropy loss is employed to learn the pixel-wise class probabilities, which is obtained by averaging the loss over the pixels with semantic labels in each mini-batch during the training phase.

\begin{equation}\label{eq:loss_semantic}
    L_{semantic} = -\frac{1}{N}\sum\limits_{i=1}^{N}c^{*}_{i}\log (c_{i})
\end{equation}
where $c_{i}=e^{z_{i}}/\sum_{c}e^{z_{i,c}}$ is the class prediction at pixel $i$ given the output $z$ of the final feature maps, $c^{*}_{i}$ is the corresponding ground truth, and $N$ is the number of pixels.

\textbf{Depth estimation.} $L_{1}$ loss is employed to learn the pixel-wise depth, which minimizes the absolute Euclidean distance between the depth prediction and the corresponding ground truth.

\begin{equation}\label{eq:loss_depth}
    L_{depth} = \frac{1}{N}\sum\limits_{i=1}^{N}|y_{i} - y^{*}_{i}|
\end{equation}
where $y_{i}$ is the depth prediction of the $i$-$th$ pixel, $y^{*}_{i}$ is the corresponding ground truth, and $N$ is the number of valid pixels.

\section{Experimental Analysis}
To demonstrate the effectiveness of the proposed SOSD-Net for simultaneous learning the monocular depth and semantic segmentation, we carry out comprehensive experiments on two publicly available datasets: CityScapes~\cite{cordts2016cityscapes} and NYU v2~\cite{silberman2012indoor}. In the following subsections, we report the details of our implementation and the evaluation results. Some ablations studies based on CityScapes are discussed to give a more detailed analysis of our method.

\subsection{Experimental Setup}
\textbf{Datasets and Data Augmentation}. The~\textbf{CityScapes} dataset~\cite{cordts2016cityscapes} is a large dataset for road scene understanding. It comprises stereo imagery from automotive-grade stereo cameras with $22cm$ baseline, labeled with instance and semantic segmentation of 20 classes. Inverse depth images are provided, labeled with the SGM method~\cite{hirschmuller2008stereo}. The dataset was collected over 50 different cities spanning several months, which consists of training, validation, and test sets containing 2,975, 500, and 1,525 images, respectively.

Following the suggestion in Ozan~\etal~\cite{sener2018multi}, the input images and the corresponding depth maps are resized to $256\times512$. The training data are augmented on the fly during the training phase. The RGB and depth images are scaled with a randomly selected ratio from $\{0.5, 0.75, 1, 1.25, 1.5, 1.75\}$. In addition, the RGB-D images are also transformed using color transformations and flip with a chance of 0.5. Please note that the proposed method has the potential to support training on full resolution input by the use of online data preparation. For example, before feeding the data to the model, we can randomly crop the input image to small patches, which will consume the same memory of GPU with the strategy of resizing input samples. However, the state-of-the-art approaches adopted the samples with a fixed resolution to train their models. For the fairness of the comparison, we employ the same resolution to train the proposed model for evaluation.

The \textbf{NYU v2}~\cite{silberman2012indoor} consists of 464 scenes ($480\times640$), captured using Microsoft Kinect. Following the official split, the training dataset is composed of 249 scenes with 795 pair-wise images, and the testing dataset includes 215 scenes with 654 pair-wise images. The input images and the corresponding depths are augmented on the fly during the training phase, which is scaled with a randomly selected ratio from $\{1, 1.2, 1.5\}$, transformed using color transformations, and flipped with a chance of 0.5.

\textbf{Evaluation Metrics.} For quantitative evaluation of the depth estimation on the NYU v2 dataset, we report errors obtained with the following widely adopted error metrics. To evaluate the performance of the semantic segmentation on the NYU v2 dataset, we use mean Intersection over Union (mIoU), mean accuracy, and pixel accuracy as metrics.
\begin{itemize}
\item
{Average relative error:
$ {\bf rel} = \frac{1}{N}\sum_{y_{i}\in|N|}\frac{|y_{i}-y^*_{i}|}{y^*_{i}} $}
\item
{Root mean squared error:
$ {\bf rms} = \sqrt{\frac{1}{N}\sum_{y_{i}\in|N|}|y_{i}-y^*_{i}|^{2}} $}
\item
{ Average $ log_{10} $ error: $ {\bf log}_{10} = \frac{1}{N}\sum_{y_{i}\in|N|}|log_{10}(y_{i})-log_{10}(y^*_{i})| $}
\item
{ Accuracy with threshold $ t $: percentage (\%) of $ y_{i} $ subject to $ max(\frac{y^*_{i}}{y_{i}}, \frac{y_{i}}{y^*_{i}})=\delta < t(t\in[1.25, 1.25^2, 1.25^3])$}
\end{itemize}
where $ y_{i} $ is the estimated depth, $ y^*_{i} $ denotes the corresponding ground truth, and $ N $ is the total number of valid pixels in all images of the validation set.

For the CityScapes dataset, we use mean absolute error and mIoU to evaluate the depth estimation and semantic segmentation, respectively.

\textbf{Implementation Details.} We implement the proposed model using both PaddlePaddle~\cite{baidupaddle} and TensorFlow frameworks, and train the network on the NVIDIA Tesla P40 with 24GB memory. The results in this paper are from the TensorFlow implementation. The objective function is optimized using Adam method~\cite{kingma2014adam}. During the initialization stage, the weight layers in the first part of the architecture are initialized using the corresponding pre-trained model (Xception) on the ILSVRC~\cite{russakovsky2015imagenet} dataset for image classification. The weights of the specific task are assigned by sampling a Gaussian with zero mean and 0.01 variance, and the learning rate is set at 0.0001. We set the batch size of the two datasets to 16. Finally, our model is trained with 60 epochs for the NYU Depth v2 dataset, and 40 epochs for the CityScapes dataset. 

\subsection{Ablation Study}
We conduct various ablation studies to analyze the performance of our approach. The baseline model (MTL) is a classical multi-task model with a backbone extracting common features, with two task-specific paths to infer the depth and the semantic, respectively, and the corresponding optimization objective is a linear combination of each branch loss. According to the description in section~\ref{sec:method}, the improved versions of the baseline include: (i) SOSD-Net (SOSD-Net: adding semantic objectness to the baseline model), (ii) ESOSD-Net (SOSD-Net with the EM learning strategy). The comparative experimental results are shown in Table~\ref{table:cityscapes_ablation}, Table~\ref{table:nyu_ablation_study_semantic} and Table~\ref{table:nyu_ablation_study_depth}. Note that the baseline model, the improved versions, and the single-task model share the same advanced backbone (Xception), which extracts the features for the subnets to infer specific-task information.

\begin{table}
\newcommand{\tabincell}[2]{\begin{tabular}{@{}#1@{}}#2\end{tabular}}
\begin{center}
\begin{tabular}{|l|c|c|c|c|}
\hline
Method & \tabincell{c}{Segmentation\\ mIoU $[\%]$} & \tabincell{c}{Disparity\\ error $[px]$} & \tabincell{c}{Inference\\ speed (ms)} & \tabincell{c}{Number of\\ parameters (M)} \\
\hline\hline
Semantic only & 62.0 & - & 139.1 & 23.4 \\
Depth only & - & 2.47 & 140.7 & 23.4 \\
MTL & 65.6 & 2.64 & 142.2 & 23.6 \\
SOSD-Net & 67.2 & 2.58 & 159.0 & 24.0 \\
ESOSD-Net & \textbf{68.2} & \textbf{2.41} & 159.0 & 24.0 \\
\hline
\end{tabular}
\end{center}
\caption{Quantitative improvement when learning semantic segmentation and depth with the proposed SOSE-Net and EM-style learning strategy. Experiments were conducted on the CityScapes dataset (sub-sampled to a resolution of $256\times512$). Results are shown from the validation set. It is clear that the inference speed and the number of parameters are comparable,  we observe an improvement of performance when training with SOSD-Net, over both single-task models and MTL. Additionally, we observe a larger improvement when training on the two-tasks using the EM-style strategy (ESOSD-Net). The result shows that SOSD-Net can automatically build a better relation to embedding the scene parsing and depth estimation, and the EM-style can learn the two-tasks more effectively.}
\label{table:cityscapes_ablation}
\end{table}

\begin{table}
\newcommand{\tabincell}[2]{\begin{tabular}{@{}#1@{}}#2\end{tabular}}
\begin{center}
\begin{tabular}{|l|c|c|c|c|c|}
\hline
Method & \tabincell{c}{Mean\\ IoU} & \tabincell{c}{ Mean\\ Accuracy} & \tabincell{c}{Pixel\\ Accuracy} \\
\hline\hline
Semanitc only & 0.385 & 0.591 & 0.687 \\
MTL & 0.417 & 0.610 & 0.710 \\
SOSD-Net & 0.433 & 0.625 & 0.722 \\
ESOSD-Net & \textbf{0.450} & \textbf{0.647} & \textbf{0.733}  \\
\hline
\end{tabular}
\end{center}
\caption{Quantitative improvement when learning semantic segmentation with our proposed model. Experiments are conducted on the NYU dataset ($480\times640$). Results are shown from the test set. It is observed that SOSD-Net with EM-style achieves better performance over both single-task models and MTL.}
\label{table:nyu_ablation_study_semantic}
\end{table}

\subsubsection{SOSD-Net}
For the CityScapes dataset, it can be observed that SOSD-Net obtains better performance than the MTL model, e.g., SOSD-Net improves the Segmentation mIoU by 1.6\% (from 65.6\% to 67.2\%), and reduces the Disparity error by 0.06 (from 2.64 to 2.58), as reported in Table~\ref{table:cityscapes_ablation}. SOSD-Net also outperforms the Semantic only method, improving the mIoU of semantic segmentation by a margin of $5.2$, while shows comparable performance with Depth only approach. In addition, comparing with independent models and MTL model, SOSD-Net still maintains comparable inference time and the number of parameters. 

Meanwhile, we also evaluated SOSD-Net model on the NYU v2 dataset.  As reported in Table~\ref{table:nyu_ablation_study_semantic}, SOSD-Net outperforms MTL (0.433 vs. 0.417, 0.625 vs. 0.610, 0.722 vs. 0.710) and Semantic only model (0.433 vs. 0.385, 0.625 vs. 0.591, 0.722 vs. 0.687) on all metrics of semantic segmentation. As for depth estimation evaluation, SOSD-Net also obtains obvious performance gains on all metrics, as shown in Table~\ref{table:nyu_ablation_study_depth}. These ablation studies demonstrate that using the semantic objectness is able to improve the performance of monocular depth estimation and semantic segmentation.

\begin{figure}[htb]
\begin{center}
   \includegraphics[width=0.76\linewidth]{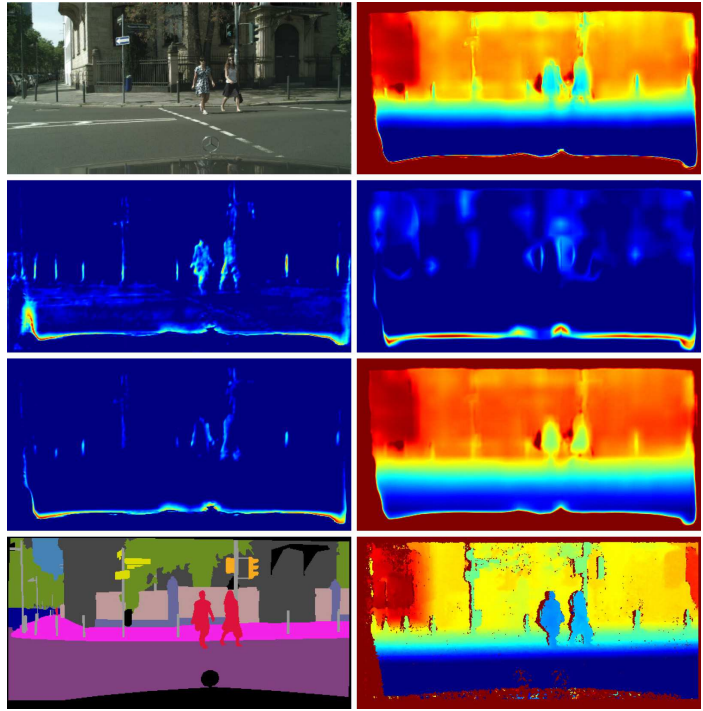}
\end{center}
   \caption{Visualization of the feature maps from the semantic-to-depth unit. The first row shows the input image and monocular depth estimation, the second row shows the feature map of the 2D latent shared representation and 3D latent shared representation, related to $(\triangle u \triangle v)^{-1}$ and $\triangle X \triangle Y$, the third row shows the depth intensity from the semantic-to-depth and pure depth branch, and the last row shows the ground truth of the semantic segmentation and depth estimation, respectively.}
\label{fig:semantic-to-depth}
\end{figure}

To further investigate the effect of SOSD-Net on the two tasks, we visualize the feature maps leaned from the semantic-to-depth unit, as shown in Figure~\ref{fig:semantic-to-depth}. The final depth is learned by fusing the information from the pure depth branch and the semantic-to-depth branch, respectively. Compared with the pure depth branch (third row, second column), the final depth (first row, second column) has a more detailed structure over the entire area of the pedestrians, which benefits from the semantic-to-depth branch (third row, first column). The visualization further verifies the outstanding performance of the semantic-to-depth branch, so it is consistent with the geometric constraint, as described in equation~\eqref{eq:area}.
\begin{table}
\newcommand{\tabincell}[2]{\begin{tabular}{@{}#1@{}}#2\end{tabular}}
\begin{center}
\begin{tabular}{|l|c|c|c|c|c|c|c|c|c|}
\hline
Method & \tabincell{c}{\bf rel} & \tabincell{c}{\bf rms} & \tabincell{c}{\bf $log_{10}$} & \tabincell{c}{\bf $\delta_{1}$} & \tabincell{c}{\bf $\delta_{2}$} & \tabincell{c}{\bf $\delta_{3}$}\\
\hline\hline
Depth only & 0.167 & 0.637 & 0.078 & 0.713 & 0.935 & 0.984 \\
MTL & 0.159 & 0.567 & 0.067 & 0.775 & 0.949 & 0.986 \\
SOSD-Net & 0.149 & 0.527 & 0.064 & 0.797 & 0.957 & 0.991 \\
ESOSD-Net & \textbf{0.145} & \textbf{0.514} & \textbf{0.062} & \textbf{0.805} & \textbf{0.962} & \textbf{0.992} \\
\hline
\end{tabular}
\end{center}
\caption{Quantitative improvement when learning monocular depth with our proposed model. Experiments are conducted on the NYU dataset ($480\times640$). Results are shown from the test set. We observe a significant performance improvement when training SOSD-Net with EM-style strategy, over both single-task models and MTL.}
\label{table:nyu_ablation_study_depth}
\end{table}

\begin{table}[htb]
\newcommand{\tabincell}[2]{\begin{tabular}{@{}#1@{}}#2\end{tabular}}
\begin{center}
\begin{tabular}{|l|c|c|}
\hline
Method & \tabincell{c}{Segmentation\\ mIoU $[\%]$} & \tabincell{c}{Disparity\\ error $[px]$} \\
\hline\hline
Kendall~\cite{kendall2018multi} & 64.2 & 2.65 \\
GradNorm~\cite{chen2018gradnorm} & 64.8 & 2.57\\
Ozan~\cite{sener2018multi} & 66.6 & 2.54\\
\hline
ESOSD-Net & \textbf{68.2} & \textbf{2.41} \\
\hline
\end{tabular}
\end{center}
\caption{Performance of the multi-task algorithms in semantic segmentation and depth estimation on the CityScapes dataset (sub-sampled to a resolution of $256\times512$).}
\label{table:cityscapes_comparison}
\end{table}

\subsubsection{EM Learning Strategy}
We verify the effectiveness of the EM learning strategy in boosting the performance of the monocular depth estimation and semantic segmentation. For the CityScapes dataset, it can be observed that the ESOSD-Net clearly outperforms MTL and SOSD-Net in the two tasks, as reported in Table~\ref{table:cityscapes_ablation}. For example, compared with the SOSD-Net, ESOSD-Net improves the Segmentation mIoU by 1.0\% (from 67.2\% to 68.2\%) and reduces the Disparity error by 0.17 (from 2.58 to 2.41). Note that in terms of inference time and the number of parameters, ESOSD-Net is the same as SOSD-Net. In addition, ESOSD-Net also outperforms the single-task models, improving the mIoU of semantic segmentation by a margin of $6.2$, and reducing the Disparity error by 0.06 (from 2.47 to 2.41).

In addition, we also evaluated ESOSD-Net on the NYU v2 dataset.  As reported in Table~\ref{table:nyu_ablation_study_semantic}, ESOSD-Net obviously outperforms Semantic only model, MTL, and SOSD-Net on all metrics of semantic segmentation. For example, compared with SOSD-Net, ESOSD-Net improves the Mean IoU by 1.7\% (from 0.433 to 0.450), the Mean Accuracy by 2.2\% (from 0.625 to 0.647), and Pixel Accuracy by 1.1\% (from 0.722 o 0.733).

As for the depth estimation evaluation, ESOSD-Net also leads to a large improvement on all metrics, as shown in Table~\ref{table:nyu_ablation_study_depth}. For example, compared with SOSD-Net, ESOSD-Net reduces the rel, rms, and $\log_{10}$ by 0.4\%, 0.013 and 0.002, and simultaneously improving the $\delta_{1}$, $\delta_{2}$, and $\delta_{3}$ by 0.8\%, 0.5\%, and 0.1\%, respectively. In addition, ESOSD-Net also outperforms the Depth only method, reducing the rel, rms, and $\log_{10}$ by 2.2\%, 0.123, and 0.016, and simultaneously improving the $\delta_{1}$, $\delta_{2}$, and $\delta_{3}$ by a margin of 9.2\%, 2.7\%, and 0.8\%, respectively. The results of the ESOSD-Net clearly outperforms single-task, MTL, and SOSD-Net, further demonstrating the effectiveness of the proposed EM learning strategy.

\subsection{Benchmark Performance}
In the first series of experiments, we focus on the CityScapes dataset~\cite{cordts2016cityscapes}. The proposed model is evaluated and compared with the state-of-the-art methods including Kendall~\etal~\cite{kendall2018multi}, GradNorm~\cite{chen2018gradnorm} and Ozan~\cite{sener2018multi}, as reported in Table~\ref{table:cityscapes_comparison}. It can be seen that our ESOSD-Net improves the accuracy of semantic segmentation by a margin of $2\%\sim4\%$, compared with previous methods in all settings. For inverse depth estimation, our ESOSD-Net outperforms the previous methods with $0.1\sim0.2$ points gap on the mean absolute error.

\begin{figure}
	\begin{center}
		\includegraphics[width=0.92\linewidth]{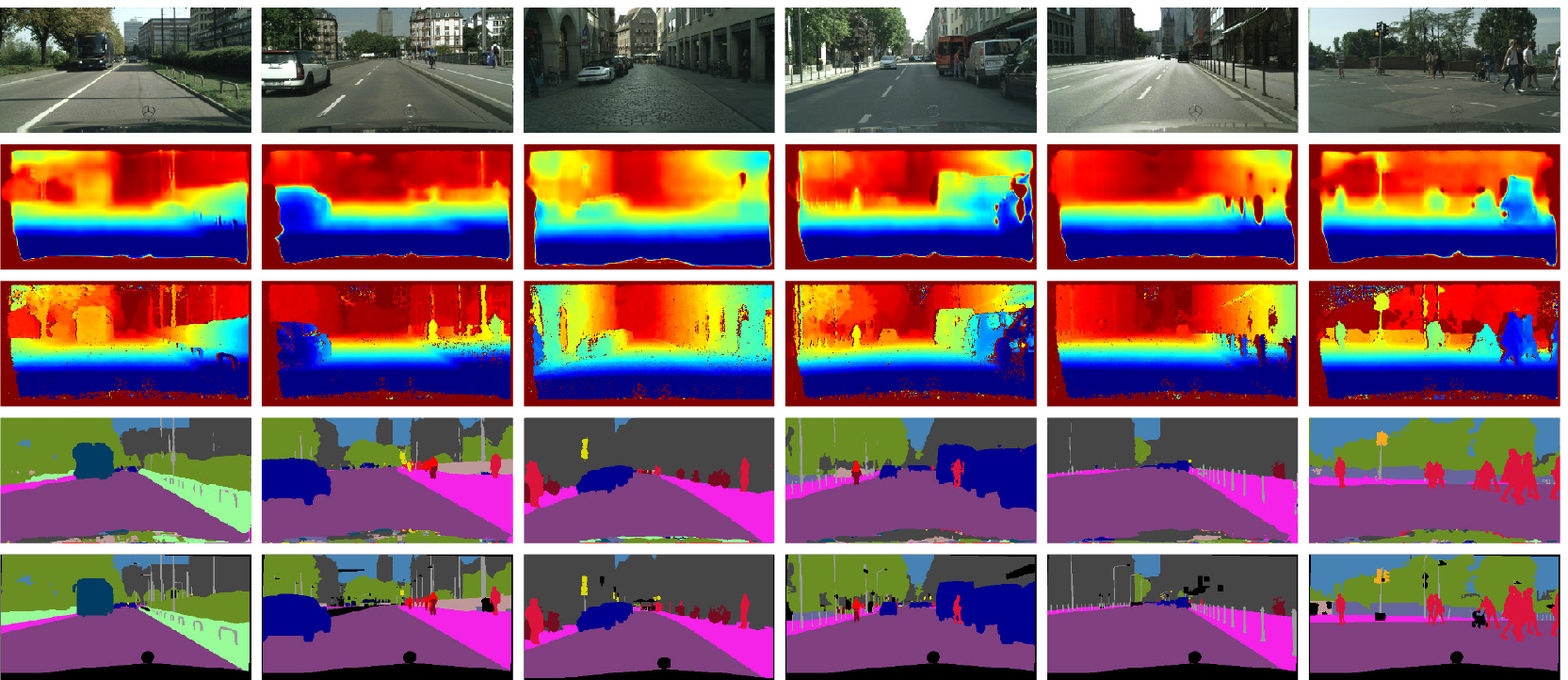}
	\end{center}
	\caption{Qualitative examples of monocular depth estimation and 19-class scene parsing results on the CityScapes dataset ($256\times512$). The second and the fourth rows corresponding to the predictions of the depth estimation and semantic segmentation. The third and the last rows corresponding to the ground truth of the depth estimation and semantic segmentation, respectively.}
	\label{fig:visualization_cityscapes}
\end{figure} 

Meanwhile, we also evaluated the proposed model on the NYU Depth v2 dataset. The comparison with the state-of-the-art algorithms are shown in Table~\ref{table:nyu_semantic_comparison} and Table~\ref{table:nyu_depth_comparison}, respectively.  As observed from Table~\ref{table:nyu_semantic_comparison}, compared  with   Deng~\etal~\cite{deng2015semantic} ,  FCN~\cite{long2015fully} ,  Eigen and Fergus~\cite{eigen2015predicting}, and Context~\cite{lin2016efficient} , the  proposed ESOSD-Net achieves a remarkable improvement. When comparing the RefineNet~\cite{lin2017refinenet}, our proposed method shows outstanding performance on the Mean Accuracy and achieves competing performance on the mean IoU and pixel accuracy. As observed in Table~\ref{table:nyu_semantic_comparison}, our proposed method is also competitive with the two-stage approaches, e.g., ESOSD-Net is comparable to the two-stage approach PAD-Net~\cite{xu2018pad} (0.450 vs. 0.502, 0.647 vs. 0.623, 0.733 vs. 0.752), and outperforms Gupta~\etal~\cite{gupta2014learning} and Arsalan~\etal~\cite{mousavian2016joint} in all metrics. These results further demonstrate the effectiveness of ESOSD-Net.

Table~\ref{table:nyu_depth_comparison} shows the evaluation result of the depth estimation. With the same number of samples (795) and a one-stage training strategy, the proposed ESOSD-Net model outperforms the state-of-the-art methods. For example, compared with the E. and F.~\cite{eigen2015predicting}, ESOSD-Net improves the rel by 1.3\% (from 0.158 to 0.145), the $\delta_{1}$ by 3.6 \% (from 76.9\% to 80.5\%), the $\delta_{2}$ by 1.2 \% (from 95.0\% to 96.2\%), and the $\delta_{3}$ by 0.4 \% (from 98.8\% to 99.2\%), respectively. Meanwhile, ESOSD-Net reports the rms of 0.514, an improvement of 0.127 over 0.641 achieved by the E. and F.~\cite{eigen2015predicting}. The experiments demonstrate the superior performance of the ESOSD-Net in depth estimation.
\begin{table}[htb]
\newcommand{\tabincell}[2]{\begin{tabular}{@{}#1@{}}#2\end{tabular}}
\begin{center}
\begin{tabular}{|l|c|c|c|}
\hline
Method & \tabincell{c}{Mean\\ IoU} & \tabincell{c}{Mean\\ Accuracy} & \tabincell{c}{Pixel\\ Accuracy} \\
\hline\hline
\bf {Two-stage:} &  \quad & \quad  & \quad  \\
Gupta~\etal~\cite{gupta2014learning} & 0.286 & - & 0.603 \\
Arsalan~\etal~\cite{mousavian2016joint} & 0.392 & 0.523 & 0.686 \\
PAD-Net~\cite{xu2018pad} & \textbf{0.502} & 0.623 & \textbf{0.752} \\
\hline
\bf {One-stage:} &  \quad & \quad  & \quad  \\
Deng~\etal~\cite{deng2015semantic} & - & 0.315 & 0.638 \\
FCN~\cite{long2015fully} & 0.292 & 0.422 & 0.600 \\
Eigen and Fergus~\cite{eigen2015predicting} & 0.341 & 0.451 & 0.656 \\
Context~\cite{lin2016efficient} & 0.406 & 0.536 & 0.700 \\
RefineNet~\cite{lin2017refinenet} & 0.465 & 0.589 & 0.736 \\
\hline
ESOSD-Net & 0.450 & \textbf{0.647} & 0.733 \\
\hline
\end{tabular}
\end{center}
\caption{Quantitative comparison with state-of-the-art methods on the scene parsing task on the NYU Depth v2 dataset ($480\times640$).}
\label{table:nyu_semantic_comparison}
\end{table}

\begin{table}[htbp]
\newcommand{\tabincell}[2]{\begin{tabular}{@{}#1@{}}#2\end{tabular}}
\begin{center}
\begin{tabular}{|l|c|c|c|c|c|c|c|}
\hline
Method & samples & {\bf rel} & {\bf rms } & ${\bf log_{10}}$ & ${\bf \delta_{1}}$ & ${\bf \delta_{2}}$ & ${\bf \delta_{3}}$ \\
\hline\hline
\bf {Two-stage:} &  \quad & \quad  & \quad  &  \quad & \quad  & \quad & \quad \\
Joint HCRF~\cite{wang2015towards} & 795 & 0.220 & 0.745 & 0.094 & 0.605 & 0.890 & 0.970 \\
Jafari~\etal~\cite{jafari2017analyzing} & 795 & 0.157  & 0.673 & 0.068 & 0.762 & 0.948 & 0.988 \\
PAD-Net~\cite{xu2018pad} & 795 & 0.120 & 0.582 & 0.055 & 0.817 & 0.954 & 0.987 \\
\hline
\bf {One-stage:} &  \quad & \quad  & \quad  &  \quad & \quad  & \quad & \quad \\
Make3D~\cite{saxena2009make3d} & 795 & 0.349 & 1.214 & - & 0.447 & 0.745 & 0.897 \\
DepthTransfer~\cite{karsch2014depth} & 795 & 0.35 & 1.20 & 0.131 & - & - & - \\
Liu~\etal~\cite{liu2014discrete} & 795 & 0.335 & 1.06 & 0.127 & - & - & - \\
Li~\etal~\cite{li2015depth} & 795 & 0.232 & 0.821 & 0.094 & - & - & - \\
Liu~\etal~\cite{liu2015deep} & 795 & 0.230 & 0.824 & 0.095 & 0.614 & 0.883 & 0.975 \\
Wang~\etal~\cite{wang2015towards} & 795 & 0.220 & 0.745 & 0.094 & 0.605 & 0.890 & 0.970 \\
Eigen~\etal\cite{eigen2014depth} & 120k & 0.215 & 0.907 & - & 0.611 & 0.887 & 0.971 \\
R. and T.~\cite{roy2016monocular} & 795 & 0.187 & 0.744 & 0.078 & - & - & - \\
E. and F.~\cite{eigen2015predicting} & 795 & 0.158 & 0.641 & - & 0.769 & 0.950 & 0.988 \\
He~\etal~\cite{he2018learning} & 48k & 0.151 & 0.572 & 0.064 & 0.789 & 0.948 & 0.98 \\
Lai~\cite{laina2016deeper} & 96k & 0.129 & 0.583 & 0.056 & 0.811 & 0.953 & 0.988 \\
DORN~\cite{fu2018deep} & 120k & \textbf{0.115} & \textbf{0.509} & \textbf{0.051} & \textbf{0.828} & \textbf{0.965} & \textbf{0.992} \\
\hline
ESOSD-Net & 795 & 0.145 & 0.514 & 0.062 & 0.805 & 0.962 & \textbf{0.992} \\
\hline
\end{tabular}
\end{center}
\caption{Quantitative comparison with state-of-the-art methods on the depth estimation task on the NYU Depth v2 dataset ($480\times640$).}
\label{table:nyu_depth_comparison}
\end{table}

When comparing with other one-stage approaches using a large number of training samples, ESOSD-Net also achieves competing performance. As reported in Table~\ref{table:nyu_depth_comparison}, ESOSD-Net outperforms He~\etal~\cite{he2018learning} on all metrics, and achieves comparable performance with Lai~\cite{laina2016deeper} and DORN~\cite{fu2018deep} (rms, $\delta_{1}$, $\delta_{2}$, $\delta_{3}$), while shows slightly weakness on rel and $\log_{10}$. This is mainly because that Lai~\cite{laina2016deeper} and DORN~\cite{fu2018deep} utilized a large number of samples: 96k and 120k, respectively, which are 120$\times$ and 150$\times$ than 795 samples used in our model.

At last, we can observe that the depth performance of ESOSD-Net is also competitive with the two-stage approaches, e.g., ESOSD-Net outperforms the two-stage approach 
Joint HCRF~\cite{wang2015towards} and Jafari~\etal~\cite{jafari2017analyzing} on all metrics. Compared with PAD-Net~\cite{xu2018pad}, ESOSD-Net achieves outstanding performance in the terms of the rms (0.514 vs. 0.582), $\delta_{2}$ (0.962 vs. 0.954) and $\delta_{3}$ (0.992 vs. 0.987), while shows slightly weakness on the rel (0.145 vs. 0.120), $\log_{10}$ (0.062 vs. 0.055), and $\delta_{1}$ (0.805 vs. 0.817). Nevertheless, it should be mentioned that PAD-Net~\cite{xu2018pad} is trained by auxiliary tasks with two-stage strategy, benefiting from the additional supervised information.

In summary, the proposed method achieves better performance than the state-of-the-art methods using a one-stage training strategy, under the same number of training samples. Furthermore, compared with the two-stage methods, the proposed ESOSD-Net also achieves competing performance. The experiments strongly demonstrate the effectiveness and superior performance of the ESOSD-Net.

\begin{figure}
\begin{center}
    \includegraphics[width=0.92\linewidth]{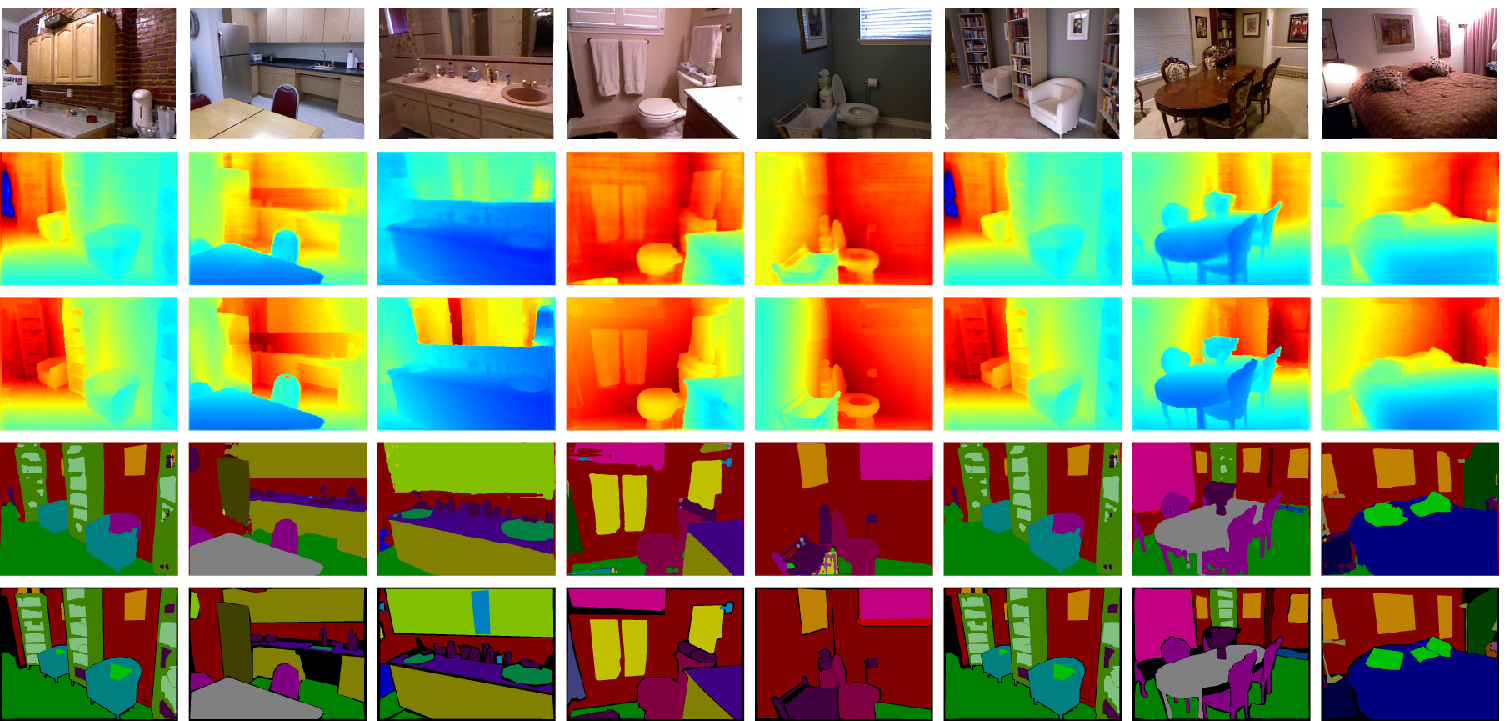}
\end{center}
   \caption{Qualitative examples of monocular depth estimation and 40-classes scene parsing results on the NYU Depth v2 dataset ($480\times640$). The second and the fourth rows corresponding to the predictions of the depth estimation and semantic segmentation. The third and the last rows corresponding to the ground truth of the depth estimation and semantic segmentation, respectively.}
\label{fig:visualization_nyu}
\end{figure}

\subsection{Visualization}
We select several samples from the CityScapes dataset and the NYU Depth v2 dataset for visualization, as shown in Figure~\ref{fig:visualization_cityscapes}. It is obvious that even there exist holes in the depth ground truth of some vehicles, ESOSD-Net can not only predict accurate depth maps with good smoothness, it can also parse the same visual semantic effect as the ground truth. In addition, as shown in Figure~\ref{fig:visualization_nyu}, the proposed model exhibits very close qualitative visualization effects in comparison with the corresponding ground truth on the NYU v2 dataset, which demonstrates the effectiveness of the proposed method.

\section{Conclusion}

In this paper, we have proposed a geometric constraint to reveal the semantic objectness relationship between the monocular depth estimation and semantic segmentation. Through this constraint, we can employ the semantic information of the scene to alleviate the ambiguity in monocular depth estimation, and simultaneously boost the accuracy of the semantic segmentation. In order to explore this constraint, the paper proposes a novel network structure (SOSD-Net) to effectively embed semantic objectness information from the geometry cues and scene parsing. We have also proposed an EM-style learning strategy to effectively train the SOSD-Net. Through extensive experimental evaluations and comparisons on the CityScapes dataset and NYU v2 dataset, the proposed ESOSD-Net achieves outstanding performance over state-of-the-art multi-task methods using the one-stage training strategy.

\section*{Acknowledgement}
This work was supported in part by the Shenzhen Fundamental Research Fund (Subject Arrangement) under Grant JCYJ20170412170602564, and the National Natural Science Foundation of China under 61822603, Grant U1813218, Grant U1713214, Grant 61672306, Grant 61572271. This work was jointly supported by Baidu Inc, Tsinghua University, and the University of Ryerson. The authors would like to thank Baidu for providing the computing resources. Thanks go to Ruijie Hou, Lixia Shen and Guangyao Yang for discussion.

\bibliography{mybibfile}

\end{document}